\documentclass[conference]{IEEEtran}
\IEEEoverridecommandlockouts
\usepackage{cite}
\usepackage{amsmath,amssymb,amsfonts}
\usepackage{algorithmic}
\usepackage{graphicx}
\usepackage{textcomp}
\usepackage{xcolor}
\def\BibTeX{{\rm B\kern-.05em{\sc i\kern-.025em b}\kern-.08em
    T\kern-.1667em\lower.7ex\hbox{E}\kern-.125emX}}
\begin{document}

\title{Multimodal Guidance Network for Missing-Modality Inference
in Content Moderation
}

\author{\IEEEauthorblockN{Zhuokai Zhao$^{*}$}\thanks{* Work done during an internship at Twitch Interactive, Inc.}
\IEEEauthorblockA{\textit{University of Chicago} \\
Chicago, USA \\
zhuokai@uchicago.edu}
\and
\IEEEauthorblockN{Harish Palani}
\IEEEauthorblockA{\textit{Twitch Interactive, Inc.} \\
San Francisco, USA \\
palanhar@twitch.tv}
\and
\IEEEauthorblockN{Tianyi Liu}
\IEEEauthorblockA{\textit{Twitch Interactive, Inc.} \\
San Francisco, USA \\
tinyli@twitch.tv}
\and
\IEEEauthorblockN{Lena Evans}
\IEEEauthorblockA{\textit{Twitch Interactive, Inc.} \\
San Francisco, USA \\
evanslen@twitch.tv}
\and
\IEEEauthorblockN{Ruth Toner}
\IEEEauthorblockA{\textit{Twitch Interactive, Inc.} \\
San Francisco, USA \\
rbtoner@twitch.tv}
\and
}

\maketitle

\begin{abstract}
Multimodal deep learning, especially vision-language models, have gained 
significant traction in recent years, greatly improving performance on many 
downstream tasks, including content moderation and violence detection. 
However, standard multimodal approaches often assume consistent modalities between
training and inference, limiting applications in many real-world use cases, as
some modalities may not be available during inference.
While existing research mitigates this problem through reconstructing the 
missing modalities, they unavoidably increase unnecessary computational cost, 
which could be just as critical, especially for large, deployed infrastructures 
in industry.
To this end, we propose a novel \textit{guidance network} that promotes knowledge 
sharing during training, taking advantage of the multimodal representations to 
train better single-modality models to be used for inference.
Real-world experiments in violence detection shows that our proposed framework 
trains single-modality models that significantly outperform traditionally 
trained counterparts, while avoiding increases in computational cost for inference.
\end{abstract}

\begin{IEEEkeywords}
Multimodal deep learning, Vision-language model, 
Content moderation, Violence detection
\end{IEEEkeywords}

\section{Introduction}\label{sec:introduction}
Multimodal deep learning~\cite{ngiam2011multimodal, zhao2024direct} has achieved great success 
and attracted tremendous attention from machine learning (ML) communities because of 
its ability to fuse representations from multiple modalities, matching how humans 
naturally perceive the world and often leading to improved performance on various 
ML tasks~\cite{baltruvsaitis2018multimodal, akkus2023multimodal}.
Despite the success that multimodal approaches have achieved in image 
classification~\cite{miller2020multi, abid2023multi, zhang2024rankclip}, video 
understanding~\cite{nagrani2020video, palaskar2022multimodal}, image 
captioning~\cite{yu2019multimodal, zhao2019multimodal, chen2024halc}, 
image-to-image translation~\cite{huang2018multimodal} and 
many others~\cite{zhao2023evaluating, chen2024autoprm},
one of the underlying assumptions is that all modalities must be consistent
between training and inference.
In other words, multimodal models are often not robust or even functional
when some of the modality used in the training stage is missing during 
inference~\cite{ma2022multimodal}.
For example,~\cite{zhu2022multimodal} requires both image and text in training 
and inference to perform video sentiment analysis.
However, such requirement greatly limits the potential of multimodal approaches 
in many application domains, such as content 
moderation~\cite{gupta2018empowering, bhandari2023crisishatemm} and violence 
detection~\cite{giannakopoulos2010multimodal, wu2020not}, where multimodal 
approaches have proven their capabilities in tremendously improving moderation 
performance, but are difficult to deploy at scale in industry due to infrastructure 
constraints and increased computational cost.
More specifically, although it is possible and often easy to collect data 
consisting of multiple modalities (such as a streamer's streaming interaction 
history in texts and contents in image and audio) during training, it is much less 
feasible and induces much more workload on the deployed moderation platform when 
collecting all these data during inference.
There are also many other reasons that can lead to inference-time missing 
modalities, including hardware-level sensor malfunctioning and
self-deficiencies~\cite{woo2023towards}.
Therefore, developing new approaches taking advantage of multiple modalities during 
training while being robust with missing modalities during inference is in 
urgent need, especially in the field of real-time content moderation and violence
detection.

Mitigating the aforementioned missing-modality problem has become a new direction
of research in the broader multimodal deep 
learning~\cite{tsai2018learning, ma2021smil} community.
Numerous existing works attempt to solve this issue through generating the missing 
modality features.
Ma et al.~\cite{ma2021smil} proposed a Bayesian Meta-Learning method named SMIL 
to estimate the latent feature of the missing-modality for data completeness.
However, such approaches suffer from only being able to work with specific 
model-modality pairs, such as ResNet~\cite{he2016deep} for images and 
LSTM~\cite{hochreiter1997long} for texts.
Ma et al.~\cite{ma2022multimodal} extended the work to more general-purposed
Transformer~\cite{vaswani2017attention} architecture for wider adaptability.
Similarly, Woo et al.~\cite{woo2023towards} proposed a modular network, 
ActionMAE, that learns the missing modality by randomly dropping modality 
features and reconstructing them later from existing modality features.

While existing works provide great insights, generating missing modality
features from existing ones itself indicates that these multimodal features 
might contain richer information than we have utilized through the traditional 
training-inference procedure.
And it triggers the question of whether we can utilize the multimodal features to 
train a better performing single-modality model for inference, so that
missing-modality is no longer a concern.
Therefore, in this paper, we propose a novel guidance network in violence 
detection that promotes knowledge sharing during training, taking advantage of 
the multimodal representations to train better single-modality model to be
used in inference.
Specifically, the proposed framework uses multimodal fusion representations
as guidance to train better single-modality encoders.
%
Compared to existing works, we are skipping the missing-feature generation step.
Instead, we explore an alternative way of utilizing multimodal representations to
improve performance of single-modal models and try to take missing modality during 
inference as an opportunity to lower the latency and cost.
\begin{figure*}[ht!]
      \centering
      \includegraphics[width=.9\textwidth]{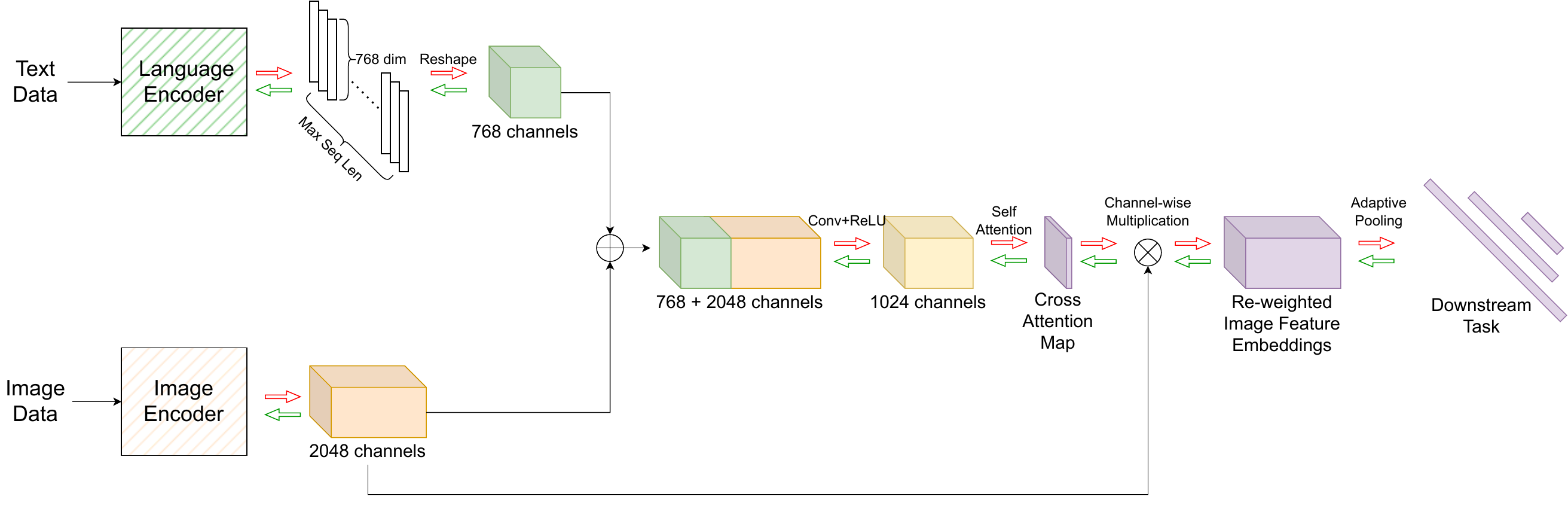}
      \caption{
      An overview of the proposed guidance network in a vision-language
      setup. The network begins with encoding text and image features separately, 
      fusing the embeddings from both modalities, and then applies self-attention 
      to obtain the attention map. However, instead of applying the attention map 
      back to the fusion embeddings, we apply it to the image-only embeddings to 
      promote knowledge sharing from cross-modality features to better singlemodal 
      attention.}
      \label{fig:train_framework}
      \vspace{-0.2in}
\end{figure*}
\section{Methodology}\label{sec:method}
%
%

%
The proposed guidance network for violence detection is illustrated in 
Fig.~\ref{fig:train_framework}.
In general, it applies a cross-modality attention map generated from fusion 
embeddings that contain both image and text features to image embedding only.
The re-weighted image embeddings are then used for training with the violence
detection task.
We illustrate the details of how we prepare the text embeddings, image 
embeddings, and how we re-weigh the image embeddings via cross-modality 
attention map in Section~\ref{subsec:method_text},~\ref{subsec:method_image}, 
and~\ref{subsec:method_reweight}, respectively.

\subsection{Text Embedding}\label{subsec:method_text}
Suppose the data we are training with are image-text pairs, the proposed
network encodes each image and its corresponding text captions through 
separate encoders.
The language encoder starts with tokenizing the input texts, pads the number of 
tokens to a maximum sequence length, and results in embeddings for each token 
with specific hidden dimension. 
Maximum sequence length and hidden dimension are two hyperparameters that are
user-defined.
In our case, unless further specified, we use 121 and 768 as our choices of 
the two parameters.
Then, we form the resulting text embeddings as a block with shape 
$\left[hidden\_dim, \sqrt{seq\_length}, \sqrt{seq\_length}\right]$.
\subsection{Image Embedding and Image-Text Fusion}\label{subsec:method_image}
For the process of preparing image features, we follow standard existing 
approaches that use feature-extracting backbones to create image embeddings, 
where the backbone is labeled as image encoder in 
Fig.~\ref{fig:train_framework}.
With the postprocessing formulation and reshaping applied to the text embeddings,
we can conveniently concatenate both modality embeddings along the hidden dimension 
axis.
A simple convolutional neural network (CNN) consisting of three convolutional layers, 
followed by ReLU~\cite{fukushima1975cognitron} and Batch 
Normalization~\cite{ioffe2015batch} layers is applied to perform the multimodal 
fusion, resulting in the 1024-channel feature block displayed in 
Fig.~\ref{fig:train_framework}.

\subsection{Text-Guided Image Embedding Re-Weighting}\label{subsec:method_reweight}
To utilize text features for improved image encoder training, we apply 
self-attention~\cite{vaswani2017attention} on the fusion embeddings to obtain the 
cross-modality attention map.
Next, instead of applying the attention map back to the fusion embedding, we 
apply it to the original image embedding before any fusion.
Then we use the attended re-weighted image embedding alone to train with specific
tasks such as image classifications.
By doing this, the knowledge we gain from text features are transferred to the
image feature domain.
The underlying reasoning behind applying such network design is to allow 
certain information within the text, such as high-level descriptions
of the stylistic traits of a group of images or specific descriptions that 
cannot be extracted from the images alone, to aid in training of the 
image backbone.

\section{Experiment}\label{sec:experiment}
We evaluate our multimodal guidance network on a real-world violence detection 
task~\cite{yao2023survey}, aiming to identify images containing violent content.
Technically speaking, violence detection is a binary classification task.
However, it presents greater difficulty due to the overlap in the visual distribution 
between realistic in-game images and actual violent events, leading to high rates of 
false-positives where non-violent in-game graphics are incorrectly classified as real 
violence.
Unfortunately, this has proven to be complex even for advanced deep learning 
models~\cite{yao2023survey}.
And in practice, live monitoring of violent streaming incidents requires an extensive 
amount of human verification to account for these high false positives.
%
%

\subsection{Data Collection}\label{subsec:experiment_data}
We collect a custom dataset from the logged historical streaming contents with 
ground truths (\texttt{violent} or \texttt{non-violent}) labeled by professionally 
trained human judges.
For each image, we formulate a corresponding text caption that contains descriptions 
consist of streamer's past streaming history as well as various metadata such as 
the streaming title, user-defined streaming category, and streaming device.
We collect around 150,000 samples in total, which are split into train and test sets 
with an 85/15 ratio.

\subsection{Baseline Approaches}\label{subsec:experiment_baselines}
\subsubsection{Contrastive Language Image Pretraining (CLIP).}\label{subsubsec:clip}
\begin{figure}
  \centering
  \includegraphics[width=.7\linewidth]{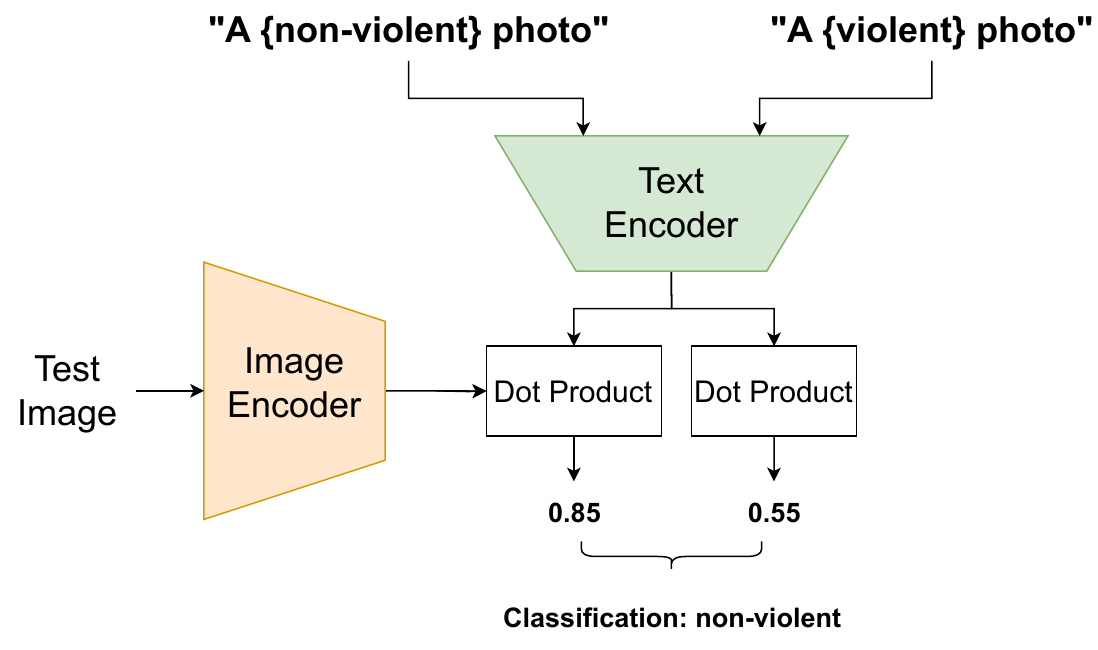}
  \caption{
    Illustration of CLIP zero-shot classification. 
    CLIP encodes both image and all potential class captions, compares similarity 
    scores between image and each text embedding and takes the higher one as the 
    classification result. 
    In the case showed here, it classifies the input image as non-violent.
  }
  \label{fig:clip_zeroshot}
  \vspace{-0.15in}
\end{figure}

The first baseline in our experiment is the pre-trained CLIP, known for its robust 
performance in various zero-shot classification tasks~\cite{radford2021learning}. 
Considering our task involves identifying violent images, which might also be present 
in other internet-sourced datasets, we anticipate that large foundation models like 
CLIP, trained on extensive internet data, should exhibit commendable performance even 
in a zero-shot scenario.
Fig.~\ref{fig:clip_zeroshot} details our approach for the baseline evaluation using 
the pre-trained CLIP.
We utilize CLIP's pre-trained image and text encoders and compare the cosine distance 
between the image and the text embeddings. 
Then the classification is determined by choosing the pair with the closer match. 
The performance tested with various image encoders pre-trained with CLIP is presented in
Table~\ref{tab:clip_performance}, with latency measurements conducted on a single 
NVIDIA V100 GPU.
Considering both performance and latency, CLIP (ViT-B/16) performs the best and 
will be used as the representative for CLIP's performance in later comparisons 
of this paper.
\begin{table}[!t]
\centering
\caption{
  Performance of the pre-trained CLIP with different vision encoders.
}
\begin{tabular}{l*{4}{c}}\hline
\textbf{Model}  &\textbf{Precision} &\textbf{Recall}    &\textbf{Accuracy}   &\textbf{Latency}        \\\hline
RN50        &94.79\%   &65.63\%   &90.32\%    &0.56ms         \\\hline
RN101       &90.63\%   &67.19\%   &89.86\%    &0.69ms         \\\hline
ViT-B/32    &\textbf{95.01}\%   &47.70\%   &86.03\%    &\textbf{0.26ms}         \\\hline
ViT-B/16    &92.94\% &\textbf{89.31\%} &\textbf{95.54\%} &0.88ms         \\\hline
ViT-L/14    &93.94\%   &85.26\%   &94.84\%    &3.39ms         \\\hline
\end{tabular}
\label{tab:clip_performance}
\vspace{-0.1in}
\end{table}

\subsubsection{MobileOne.}\label{subsubsec:mobileone}
MobileOne (MO)~\cite{vasu2023mobileone} is a state-of-the-art, efficient neural network 
backbone that has demonstrated superior performance in image classification tasks
such as ImageNet~\cite{deng2009imagenet}.
After initializing with weights from the pre-trained model, we adapt the classification
head to our dataset and fine-tuned the model specifically for our violence detection
task.
Table~\ref{tab:mobileone_performance} displays the performance of all MO variants.
As the variant number increases from MO-S0 to MO-S4, the model's complexity, number of 
trainable parameters, and performance also increase.
We refer readers to the original MO paper~\cite{vasu2023mobileone} for 
more details on these variants.
Of all the variants, MO-S4 showed the best results and will be used as 
the representative for MO performance in later comparisons.
\begin{table}[!t]
\caption{
  Fine-tuned MobileOne performance.
}
\centering
\begin{tabular}{l*{4}{c}}\hline
\textbf{Model}  &\textbf{Precision} &\textbf{Recall}    &\textbf{Accuracy}   &\textbf{Latency}        \\\hline
MO-S0      &97.27\%   &86.33\%   &95.90\%    &\textbf{0.11ms}         \\\hline
MO-S1      &\textbf{97.61}\%   &81.30\%   &94.73\%    &0.11ms         \\\hline
MO-S2      &96.16\%   &86.45\%   &95.67\%    &0.12ms         \\\hline
MO-S3      &94.33\%   &89.68\%   &95.99\%    &0.11ms         \\\hline
MO-S4      &94.05\% &\textbf{93.53\%} &\textbf{96.84\%} &0.14ms         \\\hline
\end{tabular}
\label{tab:mobileone_performance}
\vspace{-0.2in}
\end{table}

\subsection{Results of Our Guidance Approach}\label{subsec:experiment_results}
To evaluate our guidance network's ability to enhance single-modal model training using 
multimodal data, we employ MO-S4 as our image encoder, consistent with the baseline 
approach. 
For text processing, we use DistilBERT~\cite{sanh2019distilbert}, a streamlined version 
of BERT~\cite{devlin2018bert}, to handle the text captions. 
The text captions, being natural sentences, allow for flexibility in freezing or 
unfreezing backpropagation during training. 
The comparative results, including those of the two baselines, are compiled in 
Table~\ref{tab:all_performance}.
\begin{table}[!t]
\centering
\caption{
  Performance comparisons between our proposed multimodal guidance network and 
  two baselines.
}
\begin{tabular}{l*{4}{c}}\hline
\textbf{Model}  &\textbf{Precision} &\textbf{Recall}    &\textbf{Accuracy}   &\textbf{Latency}        \\\hline
CLIP        &92.94\%   &89.31\%   &95.54\%    &0.88ms         \\\hline
MO-S4       &94.05\%   &93.53\%   &96.84\%    &0.14ms         \\\hline
MO-S4 \\Frozen    &95.79\%   &\textbf{96.92}\% &98.13\% &\textbf{0.13}ms \\\hline
MO-S4 \\Unfrozen  &\textbf{97.84}\% &95.50\% &\textbf{98.32}\% &\textbf{0.13}ms   \\\hline
\end{tabular}
\label{tab:all_performance}
\vspace{-0.2in}
\end{table}

\subsection{Discussions}\label{subsec:experiment_discussion}
The guidance network we proposed excels in all metrics, including precision, recall, 
accuracy, and latency, significantly outperforming the strong baselines. 
Given that the image encoder we employ is the same as the one in MO-S4 baseline, this 
performance gain empirically validates our hypothesis that the guidance network can 
leverage multimodal representations to train a more efficient single-modal model, while 
maintaining the same network complexity and low latency. 

Interestingly, no definitive advantage is observed between guidance-trained image 
encoders when the language encoder's training status (frozen or unfrozen) varied. 
We suspect this is due to the natural-language text captions aligning with 
DistilBERT's training distribution, rendering additional fine-tuning of the 
language encoder less impactful.

\section{Conclusions and Future Work}\label{sec:conclusion}
In this paper, we propose a novel guidance network in content moderation that 
uniquely addresses the missing modality inference problem.
While current research often favors generative methods to compensate for missing 
modalities by using existing features, our paper proposes a different strategy 
focusing on harnessing the strengths of multimodal data during the training phase to 
develop a more efficient single-modal model. 
Empirical results confirm our hypothesis that our guidance network can significantly 
improve single-modal models compared to counterparts with the same architecture but 
fine-tuned traditionally, making our method balance the effectiveness of multimodal 
approaches with cost efficiency.

Inspired by our findings, we believe that it is worthwhile to continue 
researching in this direction, and could start with the following aspects.
First, better-designed attention mechanism within the guidance network 
could be studied to promote better knowledge sharing and transferring.
Next, different modality pairs, such as image-audio, or text-audio, could
be further explored for additional downstream tasks such as video understanding,
sentiment analysis, and many others.
Last but not least, we also believe that such knowledge sharing concepts could
be further extended to more than two modalities to develop even more
powerful yet efficient solutions for missing modality inference.

\bibliographystyle{IEEEbib}
\bibliography{reference}

\begin{thebibliography}{10}

\bibitem{ngiam2011multimodal}
Jiquan Ngiam, Aditya Khosla, Mingyu Kim, Juhan Nam, Honglak Lee, and Andrew~Y Ng,
\newblock ``Multimodal deep learning,''
\newblock in {\em Proceedings of the 28th international conference on machine learning (ICML-11)}, 2011, pp. 689--696.

\bibitem{zhao2024direct}
Zhuokai Zhao, Yibo Jiang, and Yuxin Chen,
\newblock ``Direct acquisition optimization for low-budget active learning,''
\newblock {\em arXiv preprint arXiv:2402.06045}, 2024.

\bibitem{baltruvsaitis2018multimodal}
Tadas Baltru{\v{s}}aitis, Chaitanya Ahuja, and Louis-Philippe Morency,
\newblock ``Multimodal machine learning: A survey and taxonomy,''
\newblock {\em IEEE transactions on pattern analysis and machine intelligence}, vol. 41, no. 2, pp. 423--443, 2018.

\bibitem{akkus2023multimodal}
Cem Akkus, Luyang Chu, Vladana Djakovic, Steffen Jauch-Walser, Philipp Koch, Giacomo Loss, Christopher Marquardt, Marco Moldovan, Nadja Sauter, Maximilian Schneider, et~al.,
\newblock ``Multimodal deep learning,''
\newblock {\em arXiv preprint arXiv:2301.04856}, 2023.

\bibitem{miller2020multi}
Stuart~J Miller, Justin Howard, Paul Adams, Mel Schwan, and Robert Slater,
\newblock ``Multi-modal classification using images and text,''
\newblock {\em SMU Data Science Review}, vol. 3, no. 3, pp. 6, 2020.

\bibitem{abid2023multi}
Muhammad~Haris Abid, Rehan Ashraf, Toqeer Mahmood, and CM~Nadeem Faisal,
\newblock ``Multi-modal medical image classification using deep residual network and genetic algorithm,''
\newblock {\em Plos one}, vol. 18, no. 6, pp. e0287786, 2023.

\bibitem{zhang2024rankclip}
Yiming Zhang, Zhuokai Zhao, Zhaorun Chen, Zhili Feng, Zenghui Ding, and Yining Sun,
\newblock ``Rankclip: Ranking-consistent language-image pretraining,''
\newblock {\em arXiv preprint arXiv:2404.09387}, 2024.

\bibitem{nagrani2020video}
Arsha Nagrani,
\newblock {\em Video understanding using multimodal deep learning},
\newblock Ph.D. thesis, University of Oxford, 2020.

\bibitem{palaskar2022multimodal}
Shruti Palaskar,
\newblock {\em Multimodal Learning from Videos: Exploring Models and Task Complexities},
\newblock Ph.D. thesis, Carnegie Mellon University, 2022.

\bibitem{yu2019multimodal}
Jun Yu, Jing Li, Zhou Yu, and Qingming Huang,
\newblock ``Multimodal transformer with multi-view visual representation for image captioning,''
\newblock {\em IEEE transactions on circuits and systems for video technology}, vol. 30, no. 12, pp. 4467--4480, 2019.

\bibitem{zhao2019multimodal}
Dexin Zhao, Zhi Chang, and Shutao Guo,
\newblock ``A multimodal fusion approach for image captioning,''
\newblock {\em Neurocomputing}, vol. 329, pp. 476--485, 2019.

\bibitem{chen2024halc}
Zhaorun Chen, Zhuokai Zhao, Hongyin Luo, Huaxiu Yao, Bo~Li, and Jiawei Zhou,
\newblock ``Halc: Object hallucination reduction via adaptive focal-contrast decoding,''
\newblock {\em arXiv preprint arXiv:2403.00425}, 2024.

\bibitem{huang2018multimodal}
Xun Huang, Ming-Yu Liu, Serge Belongie, and Jan Kautz,
\newblock ``Multimodal unsupervised image-to-image translation,''
\newblock in {\em Proceedings of the European conference on computer vision (ECCV)}, 2018, pp. 172--189.

\bibitem{zhao2023evaluating}
Zhuokai Zhao, Takumi Matsuzawa, William Irvine, Michael Maire, and Gordon~L Kindlmann,
\newblock ``Evaluating machine learning models with nero: Non-equivariance revealed on orbits,''
\newblock {\em arXiv preprint arXiv:2305.19889}, 2023.

\bibitem{chen2024autoprm}
Zhaorun Chen, Zhuokai Zhao, Zhihong Zhu, Ruiqi Zhang, Xiang Li, Bhiksha Raj, and Huaxiu Yao,
\newblock ``Autoprm: Automating procedural supervision for multi-step reasoning via controllable question decomposition,''
\newblock {\em arXiv preprint arXiv:2402.11452}, 2024.

\bibitem{ma2022multimodal}
Mengmeng Ma, Jian Ren, Long Zhao, Davide Testuggine, and Xi~Peng,
\newblock ``Are multimodal transformers robust to missing modality?,''
\newblock in {\em Proceedings of the IEEE/CVF Conference on Computer Vision and Pattern Recognition}, 2022, pp. 18177--18186.

\bibitem{zhu2022multimodal}
Tong Zhu, Leida Li, Jufeng Yang, Sicheng Zhao, Hantao Liu, and Jiansheng Qian,
\newblock ``Multimodal sentiment analysis with image-text interaction network,''
\newblock {\em IEEE Transactions on Multimedia}, 2022.

\bibitem{gupta2018empowering}
Divam Gupta, Indira Sen, Niharika Sachdeva, Ponnurangam Kumaraguru, and Arun~Balaji Buduru,
\newblock ``Empowering first responders through automated multimodal content moderation,''
\newblock in {\em 2018 IEEE International Conference on Cognitive Computing (ICCC)}. IEEE, 2018, pp. 1--8.

\bibitem{bhandari2023crisishatemm}
Aashish Bhandari, Siddhant~B Shah, Surendrabikram Thapa, Usman Naseem, and Mehwish Nasim,
\newblock ``Crisishatemm: Multimodal analysis of directed and undirected hate speech in text-embedded images from russia-ukraine conflict,''
\newblock in {\em Proceedings of the IEEE/CVF Conference on Computer Vision and Pattern Recognition}, 2023, pp. 1993--2002.

\bibitem{giannakopoulos2010multimodal}
Theodoros Giannakopoulos, Aggelos Pikrakis, and Sergios Theodoridis,
\newblock ``A multimodal approach to violence detection in video sharing sites,''
\newblock in {\em 2010 20th International Conference on Pattern Recognition}. IEEE, 2010, pp. 3244--3247.

\bibitem{wu2020not}
Peng Wu, Jing Liu, Yujia Shi, Yujia Sun, Fangtao Shao, Zhaoyang Wu, and Zhiwei Yang,
\newblock ``Not only look, but also listen: Learning multimodal violence detection under weak supervision,''
\newblock in {\em Computer Vision--ECCV 2020: 16th European Conference, Glasgow, UK, August 23--28, 2020, Proceedings, Part XXX 16}. Springer, 2020, pp. 322--339.

\bibitem{woo2023towards}
Sangmin Woo, Sumin Lee, Yeonju Park, Muhammad~Adi Nugroho, and Changick Kim,
\newblock ``Towards good practices for missing modality robust action recognition,''
\newblock in {\em Proceedings of the AAAI Conference on Artificial Intelligence}, 2023, vol.~37, pp. 2776--2784.

\bibitem{tsai2018learning}
Yao-Hung~Hubert Tsai, Paul~Pu Liang, Amir Zadeh, Louis-Philippe Morency, and Ruslan Salakhutdinov,
\newblock ``Learning factorized multimodal representations,''
\newblock {\em arXiv preprint arXiv:1806.06176}, 2018.

\bibitem{ma2021smil}
Mengmeng Ma, Jian Ren, Long Zhao, Sergey Tulyakov, Cathy Wu, and Xi~Peng,
\newblock ``Smil: Multimodal learning with severely missing modality,''
\newblock in {\em Proceedings of the AAAI Conference on Artificial Intelligence}, 2021, vol.~35, pp. 2302--2310.

\bibitem{he2016deep}
Kaiming He, Xiangyu Zhang, Shaoqing Ren, and Jian Sun,
\newblock ``Deep residual learning for image recognition,''
\newblock in {\em Proceedings of the IEEE conference on computer vision and pattern recognition}, 2016, pp. 770--778.

\bibitem{hochreiter1997long}
Sepp Hochreiter and J{\"u}rgen Schmidhuber,
\newblock ``Long short-term memory,''
\newblock {\em Neural computation}, vol. 9, no. 8, pp. 1735--1780, 1997.

\bibitem{vaswani2017attention}
Ashish Vaswani, Noam Shazeer, Niki Parmar, Jakob Uszkoreit, Llion Jones, Aidan~N Gomez, {\L}ukasz Kaiser, and Illia Polosukhin,
\newblock ``Attention is all you need,''
\newblock {\em Advances in neural information processing systems}, vol. 30, 2017.

\bibitem{fukushima1975cognitron}
Kunihiko Fukushima,
\newblock ``Cognitron: A self-organizing multilayered neural network,''
\newblock {\em Biological cybernetics}, vol. 20, no. 3-4, pp. 121--136, 1975.

\bibitem{ioffe2015batch}
Sergey Ioffe and Christian Szegedy,
\newblock ``Batch normalization: Accelerating deep network training by reducing internal covariate shift,''
\newblock in {\em International conference on machine learning}. pmlr, 2015, pp. 448--456.

\bibitem{yao2023survey}
Huiling Yao and Xing Hu,
\newblock ``A survey of video violence detection,''
\newblock {\em Cyber-Physical Systems}, vol. 9, no. 1, pp. 1--24, 2023.

\bibitem{radford2021learning}
Alec Radford, Jong~Wook Kim, Chris Hallacy, Aditya Ramesh, Gabriel Goh, Sandhini Agarwal, Girish Sastry, Amanda Askell, Pamela Mishkin, Jack Clark, et~al.,
\newblock ``Learning transferable visual models from natural language supervision,''
\newblock in {\em International conference on machine learning}. PMLR, 2021, pp. 8748--8763.

\bibitem{vasu2023mobileone}
Pavan Kumar~Anasosalu Vasu, James Gabriel, Jeff Zhu, Oncel Tuzel, and Anurag Ranjan,
\newblock ``Mobileone: An improved one millisecond mobile backbone,''
\newblock in {\em Proceedings of the IEEE/CVF Conference on Computer Vision and Pattern Recognition}, 2023, pp. 7907--7917.

\bibitem{deng2009imagenet}
Jia Deng, Wei Dong, Richard Socher, Li-Jia Li, Kai Li, and Li~Fei-Fei,
\newblock ``Imagenet: A large-scale hierarchical image database,''
\newblock in {\em 2009 IEEE conference on computer vision and pattern recognition}. Ieee, 2009, pp. 248--255.

\bibitem{sanh2019distilbert}
Victor Sanh, Lysandre Debut, Julien Chaumond, and Thomas Wolf,
\newblock ``Distilbert, a distilled version of bert: smaller, faster, cheaper and lighter,''
\newblock {\em arXiv preprint arXiv:1910.01108}, 2019.

\bibitem{devlin2018bert}
Jacob Devlin, Ming-Wei Chang, Kenton Lee, and Kristina Toutanova,
\newblock ``Bert: Pre-training of deep bidirectional transformers for language understanding,''
\newblock {\em arXiv preprint arXiv:1810.04805}, 2018.

\end{thebibliography}

\end{document}